# Conditioning Algorithms for Exact and Approximate Inference in Causal Networks

Adnan Darwiche
Rockwell Science Center
1049 Camino Dos Rios
Thousand Oaks, CA 91360
*darwiche@risc.rockwell.com*

## Abstract

We present two algorithms for exact and approximate inference in causal networks. The first algorithm, dynamic conditioning, is a refinement of cutset conditioning that has linear complexity on some networks for which cutset conditioning is exponential. The second algorithm, B-conditioning, is an algorithm for approximate inference that allows one to trade-off the quality of approximations with the computation time. We also present some experimental results illustrating the properties of the proposed algorithms.

## 1 INTRODUCTION

Cutset conditioning is one of the earliest algorithms for evaluating multiply connected networks [6]. Cutset conditioning works by reducing multiply connected networks into a number of conditioned singly connected networks, each corresponding to a particular instantiation of a loop cutset [6, 7]. Cutset conditioning is simple, but leads to an exponential number of conditioned networks. Therefore, cutset conditioning is not practical unless the size of a loop cutset is relatively small.

In this paper, we introduce the notions of relevant and local cutsets, which seem to be very effective in improving the efficiency of cutset conditioning. Relevant and local cutsets are subsets of a loop cutset [8]. We use these new notions in developing a refined algorithm, called dynamic conditioning. Dynamic conditioning has a linear computational complexity on networks such as the diamond ladder and cascaded n-bit adders, where cutset conditioning leads to an exponential behavior.

Relevant and local cutsets play the following complementary roles with respect to cutset conditioning. Relevant cutsets reduce the time required for evaluating a conditioned network using the polytree algorithm. Specifically, relevant cutsets characterize cutset variables that affect the value of each message passed by the polytree algorithm. Therefore, relevant cutsets tell us whether two conditioned networks lead to the same value of a polytree message so that the message will be computed only once. Relevant cutsets can be identified in linear time given a loop cutset and they usually lead to exponential savings when utilized by cutset conditioning.

Local cutsets, on the other hand, eliminate the need for considering an exponential number of conditioned networks. As it turns out, one need not condition on a loop cutset in order for the polytree algorithm to commence. Instead, each polytree step can be validated in a multiply connected network by only conditioning on a local cutset, which is a subset of a loop cutset. Local cutsets can be computed in polynomial time from relevant cutsets and since they eliminate the need for conditioning on a full loop cutset, they also lead to exponential savings when utilized by cutset conditioning.

Dynamic conditioning, our first algorithm in this paper, is only a refinement of cutset conditioning using the notions of local and relevant cutsets.

The second algorithm, B-conditioning, is an algorithm for approximate reasoning that combines dynamic conditioning (or another exact algorithm) with a satisfiability tester or a kappa algorithm to yield an algorithm in which one can trade the quality of approximate inference with computation time. As we shall discuss, the properties of B-conditioning depend heavily on the underlying satisfiability tester or kappa algorithm. We discuss B-conditioning and provide some experimental results to illustrate its behavior.

## 2 DYNAMIC CONDITIONING

We start this section by a review of cutset conditioning and then follow by discussing relevant and local cutsets.

### 2.1 A review of cutset conditioning

We adopt the same notation used in [6] for describing the polytree algorithm. In particular, variables are denoted by uppercase letters, sets of variables are denoted by boldface uppercase letters, and instantiations are denoted by lowercase letters. The notations $e_X^+$, $e_X^-$, $e_{UX}^+$ and $e_{XY}^-$ have the usual meanings. We also have the following definitions:

$$BEL(x) =_{def} Pr(x \wedge e)$$
$$\pi(x) =_{def} Pr(x \wedge e_X^+)$$
$$\lambda(x) =_{def} Pr(e_X^- \mid x)$$
$$\pi_X(u) =_{def} Pr(u \wedge e_{UX}^+)$$
$$\lambda_Y(x) =_{def} Pr(e_{XY}^- \mid x).$$

Following [7], we defined $BEL(x)$ as $Pr(x \wedge e)$ instead of $Pr(x \mid e)$ to avoid computing the probability of evidence $e$ when applying cutset conditioning. The polytree equations are given below for future reference:

$$BEL(x) = \pi(x)\lambda(x) \quad (1)$$
$$\pi(x) = \sum_{u_1,\ldots,u_n} Pr(x \mid u_1,\ldots,u_n) \prod_i \pi_X(u_i) \quad (2)$$
$$\lambda(x) = \prod_i \lambda_{Y_i}(x) \quad (3)$$
$$\pi_{Y_i}(x) = \pi(x) \prod_{k \neq i} \lambda_{Y_k}(x) \quad (4)$$
$$\lambda_X(u_i) = \sum_x \lambda(x) \sum_{u_k: k \neq i} Pr(x \mid u) \prod_{k \neq i} \pi_X(u_k) \quad (5)$$

The above polytree equations are valid when the network is singly connected. When the network is not singly connected, we appeal to the notion of a <u>conditioned network</u> in order to apply the polytree algorithm. Conditioning a network on some instantiation $C = c$ involves three modifications to the network. First, we remove as many outgoing arcs of $C$ as possible without destroying the connectivity of the network (<u>arc absorption</u>) [7]. Next, if the arc from $C$ to $V$ is eliminated, the probability matrix of $V$ is changed by keeping only entries that are consistent with $C = c$. Third, the instantiation $C = c$ is added as evidence to the network.

Given the notion of a conditioned network, we can now describe how cutset conditioning works. Cutset conditioning involves three major steps. First, we identify a loop cutset $\mathbf{C}$, which is a set of variables the conditioning on which leads to a singly connected network. Next, we condition the network on all possible instantiations $\mathbf{c}$ of the loop cutset and then use the polytree algorithm to compute $Pr(x \wedge e \wedge \mathbf{c})$ with respect to each conditioned network. Finally, we sum up these probabilities to obtain $Pr(x \wedge e) = BEL(x)$.

From here on, we will use the notations $\pi(x \mid \mathbf{c})$, $\lambda(x \mid \mathbf{c})$, $\pi_X(u_i \mid \mathbf{c})$, and $\lambda_{Y_i}(x \mid \mathbf{c})$ to denote the supports $\pi(x)$, $\lambda(x)$, $\pi_X(u_i)$, and $\lambda_{Y_i}(x)$ in a network that is conditioned on $\mathbf{c}$. Using this notation, cutset conditioning can be described as computing $BEL(x)$ using the sum $\sum_{\mathbf{c}} BEL(x \mid \mathbf{c})$, where $\mathbf{C}$ is a loop cutset.[1]

### 2.2 Relevant cutsets

The notion of a relevant cutset was born out of the following observations. First, the multiple applications of the polytree algorithm in the context of cutset conditioning involve many redundant computations. Second, most of this redundancy can be characterized and avoided using only a linear time preprocessing on the given network and its loop cutset. We will elaborate on these observations with an example first and then provide a more general treatment.

Consider the singly connected network in Figure 1(b), for example, which results from conditioning the multiply connected network in Figure 1(a) on a loop cutset. Assuming that all variables are binary, cutset conditioning will apply the polytree algorithm $2^{10} = 1024$ times to this network. Note, however, that when two of these applications agree on the instantiation of variables $U_1, M_3, Y_2, M_5, M_{11}$, they also agree on the value of the diagnostic support $\lambda(x)$, independently of the instantiation of other cutset variables. This means that cutset conditioning can get away with computing the diagnostic support $\lambda(x)$ only $2^5$ times. This also means that 992 of the 1024 computations performed by cutset conditioning are redundant!

The cutset variables $U_1, M_3, Y_2, M_5, M_{11}$ are called the relevant cutset for $\lambda(x)$ in this case. This relevant cutset can be identified in linear time and when taken into consideration will save 992 redundant computations of $\lambda(x)$. These savings can be achieved by storing each computed value of $\lambda(x)$ in a cache that is indexed by instantiations of relevant cutsets. When cutset conditioning attempts to compute the value of $\lambda(x)$ under some conditioning case $cc_1$, the cache is checked to see whether $\lambda(x)$ was computed before under a conditioning case $cc_2$ that agrees with $cc_1$ on the relevant cutset for $\lambda(x)$. In such case, the value of $\lambda(x)$ is retrieved and no additional computation is incurred.

More generally, each causal or diagnostic support computed by the polytree algorithm is affected by only a subset of the loop cutset, which is called its <u>relevant cutset.</u> We will use the notations $\mathbf{R}_X^+$, $\mathbf{R}_X^-$, $\mathbf{R}_{UX}^+$ and $\mathbf{R}_{XY}^-$ to denote the relevant cutsets for the supports $\pi(x)$, $\lambda(x)$, $\pi_X(u)$, and $\lambda_Y(x)$, respectively. Before we define these cutsets formally, consider the following examples of relevant cutsets in connection to Figure 1(b):

$\mathbf{R}_X^+ = U_1, N_2, N_8, N_9, N_{10}, N_{16}$ is the relevant cutset for $\pi(x)$.

---

[1] Before we end this section, we would like to stress that the notations $e_X^+$, $e_X^-$, $e_{UX}^+$ and $e_{XY}^-$ could be well defined even with respect to multiply connected networks. This means that the notations $\pi(x)$, $\lambda(x)$, $\pi_X(u_i)$ and $\lambda_{Y_i}(x)$ could also be well defined with respect to multiply connected networks. For example, the causal and diagnostic supports for variable $X$, $\pi(x)$ and $\lambda(x)$, are well defined in Figure 1(c) because the evidence decompositions $e_X^+$ and $e_X^-$ are also well defined. This observation is crucial for understanding local cutsets to be discussed in Section 2.3.

$\mathbf{R}_X^- = U_1, M_3, M_5, Y_2, M_{11}$ is the relevant cutset for $\lambda(x)$.

$\mathbf{R}_{N_5 N_4}^+ = N_8, N_9, N_{10}, N_{16}$ is the relevant cutset for $\pi_{N_4}(n_5)$.

$\mathbf{R}_{XY_1}^- = U_1, M_3, M_5$ is the relevant cutset for $\lambda_{Y_1}(x)$.

In general, a cutset variable is irrelevant to a particular message if the value of the message does not dependent on the particular instantiation of that variable.

**Definition 1 (Relevant Cutsets)** $\mathbf{R}_X^+$, $\mathbf{R}_X^-$, $\mathbf{R}_{UX}^+$ and $\mathbf{R}_{XY}^-$ *are relevant cutsets for* $\pi(x \mid \mathbf{c})$, $\lambda(x \mid \mathbf{c})$, $\pi_X(u \mid \mathbf{c})$, *and* $\lambda_Y(x \mid \mathbf{c})$, *respectively, precisely when the values of messages* $\pi(x \mid \mathbf{c})$, $\lambda(x \mid \mathbf{c})$, $\pi_X(u \mid \mathbf{c})$, *and* $\lambda_Y(x \mid \mathbf{c})$ *do not dependent on the specific instantiations of* $\mathbf{C} \setminus \mathbf{R}_X^+$, $\mathbf{C} \setminus \mathbf{R}_X^-$, $\mathbf{C} \setminus \mathbf{R}_{UX}^+$ *and* $\mathbf{C} \setminus \mathbf{R}_{XY}^-$, *respectively.*

Note that both $\{C_1, C_2, C_3\}$ and $\{C_1, C_2\}$ could be relevant cutsets for some message, according to Definition 1. This means that the instantiation of $C_3$ is irrelevant to the message. We say in this case that the relevant cutset $\{C_1, C_2\}$ is <u>tighter</u> than $\{C_1, C_2, C_3\}$.

Following is a proposal for computing relevant cutsets in time linear in the size of a network, but that is not guaranteed to compute the tightest relevant cutsets.

Let $\mathbf{A}_X^+$ denote variables that are parents of $X$ in a multiply connected network $\mathcal{M}$ but are not parents of $X$ in the network that results from conditioning $\mathcal{M}$ on a loop cutset. Moreover, let $\mathbf{A}_X^-$ denote $\{X\}$ if $X$ belongs to the loop cutset and $\emptyset$ otherwise.[2] Then (1) $\mathbf{R}_{U_iX}^+$ can be $\mathbf{A}_{U_i}^+ \cup \mathbf{A}_{U_i}^-$ union all cutset variables that are relevant to messages coming into $U_i$ except from $X$; (2) $\mathbf{R}_{XY_i}^-$ can be $\mathbf{A}_{Y_i}^+ \cup \mathbf{A}_{Y_i}^-$ union all cutset variables that are relevant to messages coming into $Y_i$ except from $X$; (3) $\mathbf{R}_X^+$ can be $\mathbf{A}_X^+$ union all cutset variables relevant to causal messages into $X$; and (4) $\mathbf{R}_X^-$ can be $\mathbf{A}_X^-$ union all cutset variables relevant to diagnostics messages coming into $X$.

As we shall see later, relevant cutsets are the key element dictating the performance of dynamic conditioning. The tighter relevant cutsets are, the better the performance of dynamic conditioning. This will be discussed further in Section 2.6.

### 2.3  Local cutsets

Relevant cutsets eliminate many redundant computations in cutset conditioning. But relevant cutsets do not change the computational complexity of cutset conditioning. That is, one still needs to consider an exponential number of conditioned networks, one for each instantiation of the loop cutset.

---

[2] If $C \in \mathbf{A}_X^+$, then the instantiation of $C$ dictates the matrix of $X$ in the conditioned network. And if $C \in \mathbf{A}_X^-$, then the instantiation of $C$ corresponds to an observation about $X$.

The notion of a local cutset addresses the above issue. We will illustrate the concept of a local cutset by an example first and then follow with a more general treatment. Consider again the multiply connected network in Figure 1(a) and suppose that we want to compute the belief in variable $X$. According to the textbook definition of cutset conditioning, one must apply the polytree algorithm to each instantiation of the cutset, which contains 10 variables in this case. This leads to $2^{10}$ applications of the polytree algorithm, assuming again that all variables are binary. Suppose, however, that we condition the network on cutset variable $U_1$, thus leading to the network in Figure 1(c). In this network, the causal and diagnostic supports for variable $X$, $\pi(x \mid u_1)$ and $\lambda(x \mid u_1)$, are well defined and can be computed independently. Moreover, the belief in variable $X$ can be computed using the polytree Equation 1:

$$BEL(x) = \sum_{u_1} \pi(x \mid u_1) \lambda(x \mid u_1).$$

Note that computing the causal support for $X$ involves a network with a cutset of 5 variables, while computing the diagnostic support for $X$ involves a network with a cutset of 4 variables. If we compute these causal and diagnostic supports using cutset conditioning, we are effectively considering only $2(2^5 + 2^4) = 96$ conditioned networks as opposed to the $2^{10} = 1024$ networks considered by cutset conditioning.

The variable $U_1$ is called a *belief cutset* for variable $X$ in this case. The reason is that although the network is not singly connected (therefore, the polytree algorithm is not applicable), conditioning on $U_1$ leads to a network in which Equation 1 is valid. In general, one does not need a singly connected network for Equation 1 to be valid. One only needs to make sure that $X$ is on every path that connects one of its descendants to one of its ancestors. But this can be guaranteed by conditioning on a local cutset:

**Definition 2 (Belief Cutset)** *A belief cutset for variable $X$, written $\mathbf{C}_X$, is a set of variables the conditioning on which makes $X$ part of every undirected path connecting one of its descendants to one of its ancestors.*

In general, by conditioning a multiply connected network on a belief cutset for variable $X$, the network becomes partitioned into two parts. The first part is connected to $X$ through its parents while the second part is connected to $X$ through its children — see Figure 1(c). This makes the evidence decompositions $\mathbf{e}_X^+$ and $\mathbf{e}_X^-$ well defined. It also makes the causal and diagnostic supports $\pi(x \mid \mathbf{c}_X)$ and $\lambda(x \mid \mathbf{c}_X)$ well defined. By appealing to belief cutsets, Equation 1 can be generalized to multiply connected networks as follows:

$$BEL(x) = \sum_{\mathbf{c}_X} \pi(x \mid \mathbf{c}_X) \lambda(x \mid \mathbf{c}_X). \qquad (6)$$

The same applies to computing the causal and diagnostic supports for a variable. Each of Equations 2

and 3 do not require a singly connected network to be valid. Instead, they only require, respectively, that (1) $X$ be on every path that goes between variables connected to $X$ through different parents, and (2) $X$ be on every path that goes between variables connected to $X$ through different children.

To satisfy these conditions, one need not condition on a loop cutset:

**Definition 3 (Causal Cutset)** *A causal cutset for variable $X$, written $\mathbf{C}_X^+$, is a set of variables such that conditioning on $\mathbf{C}_X \cup \mathbf{C}_X^+$ makes $X$ part of every undirected path that goes between variables that are connected to $X$ through different parents.*

**Definition 4 (Diagnostic Cutset)** *A diagnostic cutset for variable $X$, written $\mathbf{C}_X^-$, is a set of variables such that conditioning on $\mathbf{C}_X \cup \mathbf{C}_X^-$ makes $X$ part of every undirected path that goes between variables that are connected to $X$ through different children.*

Belief, causal, and diagnostic cutsets are what we call local cutsets.

In general, by conditioning a multiply connected network on a causal cutset for variable $X$ (after conditioning on a belief cutset for $X$), we generalize Equation 2 to multiply connected networks:

$$\pi(x \mid \mathbf{c}) = \sum_{\mathbf{c}_X^+} \sum_{u_1,\ldots,u_n} Pr(x \mid u_1,\ldots,u_n) \prod_i \pi_X(u_i \mid \mathbf{c}, \mathbf{c}_X^+). \quad (7)$$

Similarly, by conditioning a multiply connected network on a diagnostic cutset for variable $X$ (after conditioning on a belief cutset for $X$), we generalize Equation 3 to multiply connected networks:

$$\lambda(x \mid \mathbf{c}) = \sum_{\mathbf{c}_X^-} \prod_i \lambda_{Y_i}(x \mid \mathbf{c}, \mathbf{c}_X^-). \quad (8)$$

Following is an example of using diagnostic cutsets. In Figure 1(c), Equation 3 is not valid for computing the diagnostic support for variable $X$. But if we condition the network on $M_5$, thus obtaining the network in Figure 1(d), Equation 3 becomes valid. This is equivalent to using Equation 8 with $M_5$ as a diagnostic cutset for variable $X$:

$$\lambda(x \mid u_1) = \sum_{m_5} \prod_i \lambda_{Y_i}(x \mid u_1, m_5).$$

Equations 6, 7, and 8 are generalizations of their polytree counterparts. They apply to multiply connected networks as well as to singly connected ones. These equations are similar to the polytree equations except for the extra conditioning on local cutsets. Computing local cutsets is very efficient given a loop cutset, a topic that will be explored in Section 2.4. But before we end this section, we need to show how to compute the causal and diagnostic supports that variables send to their neighbors.

In the polytree algorithm, the message $\pi_{Y_i}(x)$ that variable $X$ sends to its child $Y_i$ can be computed from the causal support for $X$ and from the messages that $X$ receives from its children except child $Y_i$. In multiply connected networks, however, these supports are not well defined unless we condition on local cutsets first. That is, to compute the message $\pi_{Y_i}(x)$, we must first condition on a belief cutset for $X$ to split the network into two parts, one above and one below $X$. We must then condition on a diagnostic cutset for $X$ to split the network below $X$ into a number of sub-networks, each connected to a child of $X$. That is, the message that variable $X$ sends to its child $Y_i$ is computed as follows:

$$\pi_{Y_i}(x \mid \mathbf{c}) = \sum_{\mathbf{c}_X} \pi(x \mid \mathbf{c}, \mathbf{c}_X) \sum_{\mathbf{c}_X^-} \prod_{k \neq i} \lambda_{Y_k}(x \mid \mathbf{c}, \mathbf{c}_X, \mathbf{c}_X^-).$$
(9)

This generalizes Equation 4 to multiply connected networks.

Similarly, we compute the message $\lambda_X(u_i)$ as follows:

$$\lambda_X(u_i \mid \mathbf{c}) = \sum_{\mathbf{c}_X} \sum_{x} \lambda(x \mid \mathbf{c}, \mathbf{c}_X)$$

$$\sum_{\mathbf{c}_X^+} \sum_{u_k : k \neq i} Pr(x \mid \mathbf{u}) \prod_{k \neq i} \pi_X(u_k \mid \mathbf{c}, \mathbf{c}_X, \mathbf{c}_X^+). \quad (10)$$

This generalizes Equation 5 to multiply connected networks.

Equations 6, 7, 8, 9 & 10 are the core of the dynamic conditioning algorithm. Again, these equations parallel the ones defining the polytree algorithm [6, 7]. The only difference is the extra conditioning on local cutsets, which makes the equations applicable to multiply connected networks.

### 2.4 Relating local and relevant cutsets

What is most intriguing about local and relevant cutsets is the way they relate to each other. As we shall see, local cutsets can be computed in polynomial time from relevant cutsets and the computation has a very intuitive meaning. First, cutset variables that are relevant to both the causal support $\pi(x)$ and the diagnostic support $\lambda(x)$ constitute a belief cutset for variable $X$. Next, cutset variables that are relevant to more than two causal messages $\pi_X(u_i)$ constitute a causal cutset for variable $X$. Finally, cutset variables that are relevant to more than two diagnostic messages $\lambda_{Y_i}(x)$ constitute a diagnostic cutset for variable $X$.

**Theorem 1** *We have the following:*

1. $\mathbf{R}_X^+ \cap \mathbf{R}_X^-$ *constitutes a belief cutset for variable $X$.*

2. $\mathbf{A}_X^+ \cup \left(\bigcup_{i,j} \mathbf{R}_{U_i X}^+ \cap \mathbf{R}_{U_j X}^+\right)$ *constitutes a causal cutset for variable* $X$.

3. $\mathbf{A}_X^- \cup \left(\bigcup_{i,j} \mathbf{R}_{XY_i}^- \cap \mathbf{R}_{XY_j}^-\right)$ *constitutes a diagnostic cutset for variable* $X$.

*Here,* $i \neq j$.

Intuitively, if two computations are to be made independent, one must fix the instantiation of cutset variables that are relevant to both of them. Local cutsets attempt to make computations independent. Relevant cutsets tell us what computations depend on what variables. Hence the above relation between the two classes of cutsets.

### 2.5 Dynamic conditioning

The dynamic conditioning algorithm as described in this section is oriented towards computing the belief in a single variable. To compute the belief in every variable of a network, one must apply the algorithm to each variable individually. But since a cache is being maintained, the results of computations for one variable are utilized in the computations for another variable.

To compute the belief in variable $X$, the algorithm proceeds as follows. For each instantiation of $\mathbf{C}_X$, it computes the supports $\pi(x \mid \mathbf{c}_X)$ and $\lambda(x \mid \mathbf{c}_X)$, combines them to obtain $BEL(x \mid \mathbf{c}_X)$, and then sums the results of all instantiations to obtain $BEL(x)$. This implements Equation 6. To compute the causal support $\pi(x \mid \mathbf{c}_X)$, Equation 7 is used. And to compute the diagnostic support $\lambda(x \mid \mathbf{c}_X)$, Equation 8 is used. Applying these two equations invokes the application of Equations 9 and 10, which are used to compute the messages directed from one variable to another.

If we view the application of an equation as a request for computing some support, then computing the belief in a variable causes a chain reaction in which each request leads to a set of other requests. This sequence of requests ends at the boundaries of the network.

Therefore, the control flow in the dynamic conditioning algorithm is similar to the first pass in the revised polytree algorithm [7]. The only difference is the extra conditioning on local cutsets. For example, the causal support for variable $X$ in Figure 1(b) will be computed twice, once for each instantiation of $U_1$; and the diagnostic message for $X$ from its child $Y_1$ will be computed four times, once for each instantiation of the variables $\{U_1, M_5\}$.

To avoid redundant computations, dynamic conditioning stores the value of each computed support together with the instantiation of its relevant cutset in a cache. Whenever the support is requested again, the cache is checked to see whether the support has been computed under the same instantiation of its relevant cutset. For example, when computing the belief in $V_6$ in Figure 2, the causal support $\pi_{V_3}(v_1)$ will be requested four times, once for each instantiation of the variables in $\{V_2, V_5\}$. But the relevant cutset of this support is $\mathbf{R}_{V_1 V_3}^+ = \{V_2\}$. Therefore, two of these computations are redundant since the instantiation of variable $V_5$ is irrelevant to the value of $\pi_{V_3}(v_1)$ in this case.

To summarize, the control flow in the dynamic conditioning algorithm is similar to the first pass in the revised polytree algorithm except for the conditioning on local cutsets and for the maintenance of a cache that indexes computed supports by the instantiation of their relevant cutsets.

To give a sense of the savings that relevant and local cutsets lead to, we mention the following examples. First, to compute the belief in variable $V_6$ in the diamond ladder of Figure 2, the dynamic conditioning algorithm passes only two messages between any two variables, independently of the ladder's size. Note, however, that the performance of cutset conditioning is exponential in the ladder's size. A linear behavior is also obtained in a network structure that corresponds to an n-bit adder. Here again, cutset conditioning will lead to a behavior that is exponential in the size of the adder. Finally, in the network of Figure 1(a), which has a cutset of 10 variables, dynamic conditioning computes the belief in variable $X$ by passing at most eight messages between any two variables in the network. The textbook definition of cutset conditioning passes 1024 across each arc in this case.

### 2.6 Relevant Cutsets: The Hook for Independence

Relevant cutsets are the key element dictating the performance of dynamic conditioning. The tighter relevant cutsets are, the better the performance of dynamic conditioning. We can see this in two ways. First, tighter relevant cutsets mean smaller local cutsets and, therefore, less conditioning cases. Second, tighter relevant cutsets mean less redundant computations.

Deciding what cutsets are relevant to what messages is a matter of identifying independence. Therefore, the tightest relevant cutsets would require complete utilization of independence information, which explains the title of this section.

Therefore, if one is to compute the tightest relevant cutsets, then one must take available evidence into consideration. Evidence could be an important factor because some cutset variables may become irrelevant to some messages given certain evidence.

Most existing algorithms ignore evidence in the sense that they are justified for all patterns of evidence that might be available. This might simplify the discussion and justification of algorithms, but may also lead to unnecessary computational costs. This fact is well known in the literature and is typically handled by various optimizations that are added to algorithms or by preprocessing that prunes part of a network. Most of these pruning and optimizations are rooted in considerations of independence, but there does not seem to be a way to account for them consistently.

It is our belief that the notion of relevant cutsets is a useful start for addressing this issue. Relevant cutsets do provide a very simple mechanism for translating independence information into computational gains. We have clearly not utilized this mechanism completely in this paper, but this is the subject of our current work in which we are targeting an algorithm for computing relevant cutsets that is complete with respect to d-separation.[3]

## 3  B-CONDITIONING

B-conditioning is a method for the approximate updating of causal networks. B-conditioning is based on an intuition that has underlied formal reasoning for quite a while: "Assumptions about the world do simplify computations." The difficulty in formalizing this intuition, however, has been in (a) characterizing what assumptions are good to make and (b) utilizing these assumptions computationally.

The answer to (a) is very task dependent. What makes a good assumption in one task may be a very unwise assumption in another. But in this paper, we are only concerned with the task of updating probabilities in causal networks. In this regard, suppose that computing $Pr(x \wedge a)$ is easier than computing $Pr(x)$. Therefore, the assumption $a$ would be good from a computational viewpoint as long as $Pr(x \wedge a)$ is a good approximation of $Pr(x)$. But this would hold only if $Pr(x \wedge \neg a)$ is very small. Therefore, the value of $Pr(x \wedge \neg a)$ measures the quality of the assumption $a$ from an approximation viewpoint.[4]

The answer to (b) is clear in causal networks: we can utilize assumptions computationally by using them to instantiate variables, thus cutting out arcs, and simplifying the topology of a causal network. At one extreme, we can assume the value of each cutset variable, which would reduce a network to a polytree and make our inference polynomial. But this may not lead to a good approximation of the exact probabilities. Typically, one would instantiate some of the cutset variables, thus reducing the number of loops in a network but not eliminating them completely.

In utilizing assumptions as mentioned above, one must adjust the underlying algorithm so that it computes $Pr(x \wedge e)$ as opposed to $Pr(x \mid e)$ since $Pr(x \wedge a)$ is the approximation to $Pr(x)$ in this case, not $Pr(x \mid a)$.

Now suppose that a variable $V$ has multiple values, say three of them $v_1, v_2$ and $v_3$. Suppose further that our assumption is that $v_2$ is impossible. This assumption is typically called a "finding," as opposed to an "observation." Therefore, it cannot really help in absorbing some of the outgoing arcs from $V$. Does this mean that this assumption is not useful computationally? Not really! Whenever the algorithm sums over states of variables, we can eliminate those states that contradict with the assumption (finding).[5] This could lead to great computational savings, especially in conditioning algorithms.

These are the two ways in which assumptions are utilized computationally by B-conditioning.

**What are good assumptions?**

The question now is, How do we decide what assumptions to make? Since the quality of assumptions affect both the quality of approximation and the computation time, it would be best to allow the user to trade-off these parameters. Therefore, B-conditioning allows the user to specify a parameter $\epsilon \in (0, 1)$ and uses it as a cutoff to decide on which assumptions to make. That is, as $\epsilon$ gets smaller, fewer assumptions are made, and a better approximation is obtained, but a longer computation time is expected. As $\epsilon$ gets bigger, more assumptions are made, and a worse approximation is obtained, but the computation is faster.

The user of B-conditioning would iterate over different values of $\epsilon$, starting from large epsilon to smaller ones, or even automate this iteration through code that takes the increment for changing $\epsilon$ as a parameter.

Before we specify how $\epsilon$ is used, we mention a useful property of B-conditioning: we can judge the quality of its approximations without knowing the chosen value of $\epsilon$:[6]

$$Pr(x \wedge a) \leq Pr(x) \leq Pr(x \wedge a) + 1 - \sum_{X=y} Pr(y \wedge a).$$

That is, B-conditioning provides an upper and a lower bound on the exact probability. If these bounds are not satisfactory, the user would then choose a smaller $\epsilon$ and re-apply B-conditioning.

**From $\epsilon$ to assumptions**

The parameter $\epsilon$ is used to abstract the probabilistic causal network into a propositional database $\Delta$. In particular, for each conditional probability $Pr(c \mid d) = p$, we add to $\Delta$ the propositional formula $\neg(c \wedge d)$ iff $p \leq \epsilon$. We then make $a$ an assumption iff $\Delta \cup \{x \wedge \neg a\}$ is unsatisfiable. Intuitively, when $x \wedge \neg a$ is inconsistent with the logical abstraction of a causal network, we interpret this as meaning that the probability of $x \wedge \neg a$ is very small (relative to the choice of $\epsilon$). Note that whether $\Delta \cup \{x \wedge \neg a\}$ is unsatisfiable depends mainly on $\Delta$ which depends on both the causal network and the chosen $\epsilon$.

---

[3] Even then we would not be finished since there are independences that are uncovered by d-separation, those hidden in the probability matrices associated with variables.

[4] We can also use $Pr(\neg a)$ as the measure since $Pr(x \wedge \neg a) \leq Pr(\neg a)$, but $Pr(x \wedge \neg a)$ is more informative.

[5] In dynamic conditioning, this is implemented by simply modifying the code for summing over instantiations of local cutsets so that it ignores instantiations that contradict with the assumptions.

[6] Note that $1 - \sum_{X=y} Pr(y \wedge a)$ is $Pr(\neg a)$ which is no less than $Pr(x \wedge \neg a)$.

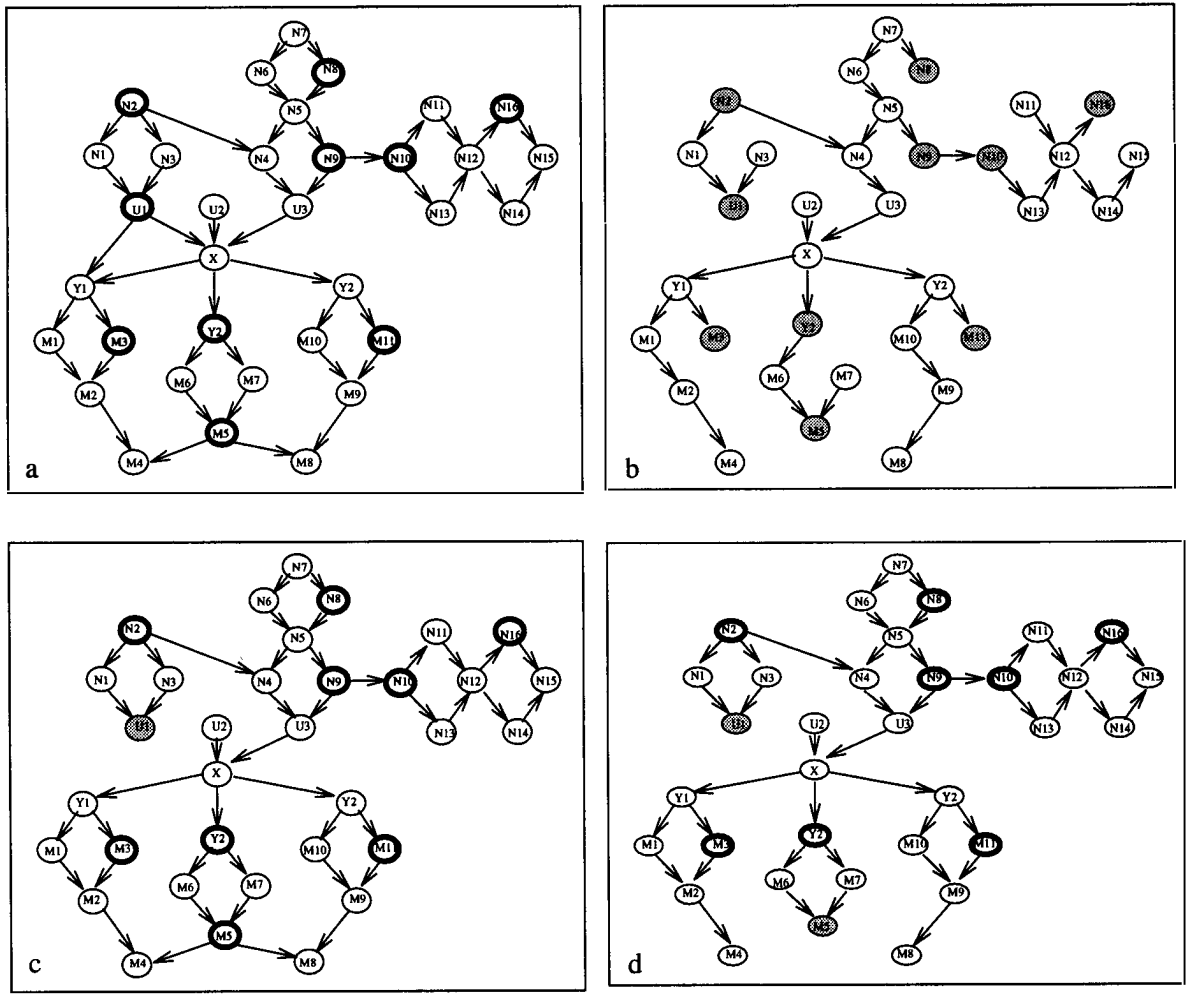

Figure 1: Example networks to illustrate global, local and relevant cutsets. Bold nodes represent a loop cutset. Shaded nodes represent cutset variables that are conditioned on.

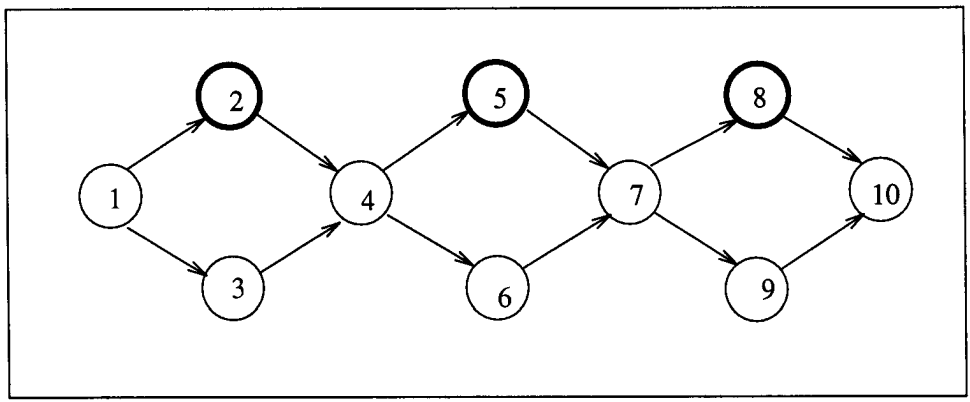

Figure 2: A causal network. Loop cutset variables are denoted by bold circles.

Alternatively, we can abstract the probabilistic network into a kappa network as suggested in [2]. We can then use a kappa algorithm to test whether $\kappa(x \wedge \neg a) > 0$. This leads to similar results since the kappa calculus is isomorphic to propositional logic in the case where all we care about is whether the kappa ranking is equal to zero.[7] As we shall see later, our implementation of B-conditioning utilizes this transformation.

### Complexity issues

If the satisfiability tester or kappa algorithm takes no time (gives almost immediate response), then B-conditioning is a good idea. But if they take more considerable time, then more issues need to be considered. But in general, we expect that the time for running satisfiability tests or kappa algorithms would be low compared to applying the exact inference algorithm. The evidence for this stems from (a) the advances that have been made on satisfiability testers recently and (b) the results on kappa algorithms as reported in [4], where a linear (but incomplete) algorithm for prediction is presented. We have used this algorithm, called k-predict, in our implementation of B-conditioning and the results were very satisfying [3]. A sample of these results are reported in the following section.[8]

### Prediction?

B-conditioning, as described here, is a method for predictive inference since we assumed no evidence (except possibly on root nodes). To handle non-predictive inference, the algorithm can be used to approximate $Pr(x \wedge e)$ and $Pr(e)$ and then use these results to approximate $Pr(x \mid e)$. But this may lead to low quality approximations. Other extensions of B-conditioning to non-predictive inference is the subject of current research.

### The choice of epsilon

Table 1 shows a number of experiments that illustrate how the value of epsilon affects the quality of approximation and time of computation.[9] The experiments concern an action network (temporal network) with 60 nodes in the domain of non-combatant evacuation [3]. Each scenario corresponds to computing the probability of successful arrival of civilians to a safe heaven given some actions (evidence) — this is a prediction task since actions are always root nodes.

A number of observations are in order about these experiments. First, a smaller epsilon may improve the quality of approximation without incurring a big computational cost. Consider the change from $\epsilon = .2$ to $\epsilon = .1$ in the first set of experiments. Here, the time of computation (in seconds) did not change, but the lower bound on the probability of unsuccessful arrival went up from .81 to .95. Note, however, that the change from $\epsilon = .1$ to $\epsilon = .02$ more than doubled the computation time, but only improved the bound with .04.

The quality of an approximation, although low, may suffice for a particular application. For example, if the probability that a plan will fail to achieve its goal is greater than .4, then one might really not care how much greater is the probability of failure [3].

The bigger the epsilon, the more the assumptions, and the lower the quality of approximations. Note, however, that some of these assumptions may not be significant computationally, that is, they do not cut any loops. Therefore, although they may degrade the quality of the approximation, they may not buy us computational time. The first two experiments illustrate this since the three additional assumptions going from $\epsilon = .1$ to $\epsilon = .2$ did not reduce computational time.

## CONCLUSION

We introduced a refinement of cutset conditioning, called dynamic conditioning, which is based on the notions of relevant and local cutsets. Relevant cutsets seem to be the critical element in dictating the computational performance of dynamic conditioning since they identify which members of a loop cutset affect the value of polytree messages. The tighter relevant cutsets are, the better the performance of dynamic conditioning. We did not show, however, how one can compute the tightest relevant cutsets in this paper.

We also introduced a method for approximate inference, called B-conditioning, which requires an exact inference method, together with either a satisfiability tester or a kappa algorithm. B-conditioning allows the user to trade-off the quality of a approximation with computational time and seems to be a practical tool as long as the satisfiability tester or the kappa algorithm has the right computational characteristics.

---

[7]The mapping is $\kappa(x) > 0$ iff $\Delta \cup \{x\}$ is unsatisfiable, where the kappa ranking and the database $\Delta$ are obtained as given above.

[8]The combination of k-predict with cutset/dynamic conditioning has been called $\epsilon$-bounded conditioning in [1].

[9]These experiments are not meant to evaluate the performance of B-conditioning, which is outside the scope of this paper. We have also eliminated experimental results that were reported in a previous version of this paper on comparing the performance of implementations of dynamic conditioning and the Jensen algorithm. Such results are hard to interpret given the vagueness in what constitutes preprocessing. In this paper, we refrain from making claims about relative computational performance and focus on stressing our contribution as a step further in making conditioning methods more competitive practically.

| $\epsilon$ | Cutset Size | Number of Assumptions | Lower Bounds [Yes/No] | Time (secs) Successful-Arrival | Time (secs) All Variables |
|---|---|---|---|---|---|
| .2 | 24 | 44 | [0/.81] | 2 | 6 |
| .1 | 24 | 41 | [0/.95] | 2 | 6 |
| .02 | 18 | 18 | [0/.99] | 5 | 21 |
| .2 | 24 | 44 | [.57/.24] | 1 | 7 |
| .1 | 24 | 41 | [.67/.29] | 2 | 6 |
| .02 | 20 | 23 | [.68/.31] | 3 | 22 |

Table 1: Experimental results for B-conditioning on a network with 60 nodes. Each set of experiments corresponds to different evidence (plan). Successful-arrival is the main query node and has two possible values, Yes and No. The table also reports the time it took to evaluate the full network (all variables). The reported lower bounds are for the successful-arrival node only. For example, [.57/.24] means that .57 is the computed lower bound for the probability of successful-arrival = Yes, while .24 is the lower bound for the probability of successful-arrival = No. The mass lost in this approximation is $1 - .57 - .24 = .19$.

The literature contains other proposals for improving the computational behavior of conditioning methods. For example, the method of bounded conditioning ranks the conditioning cases according to their probabilities and applies the polytree algorithm to the more likely cases first [5], which is closely related to B-conditioning. This leads to a flexible inference algorithm that allows for varying amounts of incompleteness under bounded resources. Another algorithm for enhancing the performance of cutset conditioning is described in [9], which also appeals to the intuition of local conditioning but seems to operationalize it in a completely different manner. The concept of knots has been suggested in [7] to partition a multiply connected network into parts containing local cutsets. This is very related to relevant cutsets, because when a message is passed between two knots, only cutset variables in the originating knot will be relevant to the message.

## Acknowledgment

I would like to thank Paul Dagum, Eric Horvitz, Moises Goldszmidt, Mark Peot, and Sampath Srinivas for various helpful discussions on the ideas in this paper. In particular, Moises Goldszmidt's work on the k-predict algorithm was a major source of insight for conceiving B-conditioning. This work has been supported in part by ARPA contract F30602-91-C-0031.